\documentclass[english]{lni}

\usepackage{graphicx}
\usepackage{fancyhdr}
\usepackage{listings} 
\usepackage{changepage} 
\usepackage[figurename=Fig., tablename=Tab., small]{caption}[2008/04/01]
\usepackage{amsmath}
\usepackage[utf8]{inputenc} 
\usepackage[T1]{fontenc}    
\usepackage{hyperref}       
\usepackage{url}            
\usepackage{booktabs}       
\usepackage{amsfonts}       
\usepackage{nicefrac}       
\usepackage{microtype}      
\usepackage{xcolor}         
\usepackage{subfiles}
\usepackage{multirow}
\usepackage{caption}
\usepackage{subcaption}
\usepackage{graphicx,wrapfig,lipsum}

\fancypagestyle{titlepage}{
\fancyhead[RO]{} 
\fancyfoot{}}

\renewcommand{\headrulewidth}{0.4pt} 

\author{Wiktor Lazarski\footnote{Tooploox, wiktor.lazarski@tooploox.com} \, Maciej Zieba\footnote{Tooploox, Wroclaw University of Science and Technology, maciej.zieba@pwr.edu.pl} \, Tanguy Jeanneau\footnote{Worldcoin, tanguy.jeanneau@worldcoin.org} \, Tobias Zillig\footnote{Worldcoin, tobias.zillig@worldcoin.org} \, Christian Brendel \footnote{Worldcoin, christian.brendel@worldcoin.org}}

\title{Two-headed eye-segmentation approach for biometric identification}

\begin{document}

\maketitle

\setcounter{footnote}{2} 
\thispagestyle{titlepage}
\pagestyle{fancy}
\fancyhead{} 
\fancyhead[RO]{\small Two-headed eye segmentation approach for biometric identification \hspace{25pt}  \hspace{0.05cm}}
\fancyhead[LE]{\hspace{0.05cm}\small  \hspace{25pt} Wiktor Lazarski, Maciej Zieba, Tanguy Jeanneau, Tobias Zillig, Christian Brendel}
\fancyfoot{} 
\renewcommand{\headrulewidth}{0.4pt} 

\begin{abstract}
Iris-based identification systems are among the most popular approaches for person identification. Such systems require good-quality segmentation modules that ideally identify the regions for different eye components. This paper introduces the new two-headed architecture, where the eye components and eyelashes are segmented using two separate decoding modules. Moreover, we investigate various training scenarios by adopting different training losses. Thanks to the two-headed approach, we were also able to examine the quality of the model with the convex prior, which enforces the convexity of the segmented shapes. We conducted extensive evaluation of various learning scenarios on real-life conditions high-resolution near-infrared iris images.


\end{abstract}

\section{Introduction}

 Biometric-based person identification systems are increasingly popular. By using a person’s intrinsic characteristics or features for identification, they remove the need for external codes, passwords, or other applications that can be easily stolen or lost. Several commonly used biometrics characteristics include fingerprints, palmprints, iris, face, and hand geometry. In this work, we are focused on iris-based identification, which extracts very discriminative features but requires an extensive processing pipeline. One of the most important steps in the pipeline is segmenting the components of the eye to locate the areas of crucial components for identification. 

Several works are focused on the problem of eye segmentation. In \cite{rot2018deep} the authors propose the simple auto-encoding architecture designed to predict both eye components and eyelashes. In \cite{liu2020convex} the authors utilize the convex prior in order to enforce the convexity of the shape. The approach described in \cite{omran2020iris} is focused entirely on the iris region during segmentation. The proposed methods are missing the extensive ablation studies investigating various loss components. 

Therefore, we propose a novel two-headed architecture that separates the segmentation of shapes of the eye and eyelashes. We investigate various types of loss components together with the convex prior approach. To summarize, our contributions are as follows. We propose a novel two-headed architecture for parallel eye components and eyelashes segmentation. We introduce complex experimental analysis that investigates various training scenarios.

\section{Related Works}
 Semantic segmentation is one of the most well-defined tasks in computer vision \cite{hao2020brief}. The most standard approaches assume access to ground truth segmentation masks and are trained in a supervised fashion. The process of annotating the images for segmentation requires assigning a label to each pixel value. As a consequence, annotating images for semantic segmentation is challenging and time-consuming. Therefore, weakly-supervised \cite{ ahn2018learning}, semi-supervised \cite{luo2020semi}, and unsupervised methods \cite{van2021unsupervised} are gaining the popularity. Unfortunately, biometric identification requires high accuracy of segmentation so the application of such methods in this domain are still limited.   

One of the first deep learning models applied to semantic segmentation was the full convolution network (FCN) proposed in \cite{long2015fully}. FCN was the first model entirely composed of convolutions that utilized upsampling using deconvolution layers in order to extract the output segmentation map. U-Net \cite{ronneberger2015u} was another breakthrough in semantic segmentation, especially widely applied in medical applications. One of the main contributions of the U-net is a specific architecture that utilizes direct connections between downsampling and upsampling paths in the model. Thanks to that approach, context information is transformed directly into higher resolution layers. PSPNet \cite{zhao2017pyramid} is another important model that introduced first the pyramid parsing module that exploits global context information by different-region based. The authors of \cite{yu2015multi} introduce DilatedNet, aggregating multi-scale contexts based on dilated convolution. DeeplabV1 \cite{chen2014semantic} started the family one of the most important segmentation models, that was further extended as  DeeplabV2\cite{chen2017deeplab}, DeeplabV3 \cite{chen2017rethinking}, and DeeplabV3+ \cite{chen2018encoder}. DeeplabV1 introduces atrous convolutions and utilizes conditional random files (CRF) to improve the localization of object boundaries. DeeplabV2 enriches atrous convolutions with atrous spatial pyramid pooling (ASPP) to robustly segment objects at multiple scales. 
DeepLabV3 Extend ASPP by introducing global pooling to the parallel branches. DeepLabv3+ extends DeepLabv3 by adding a simple yet effective decoder module to refine the segmentation results, especially along object boundaries. 
 There are also a few approaches dedicated to the problem of eye segmentation. In \cite{rot2018deep} the authors propose an encoding-decoding segmentation model that gives predictions to eye components and eyelashes. In \cite{liu2020convex} authors propose to enforce the convexity of the segmented shape by incorporating additional shapes prior to the architecture. The authors of \cite{omran2020iris} propose using a simple convolution-based model to identify only the iris region.  

\section{Our Approach}
\begin{figure}[t]
\centering
\begin{subfigure}{0.47\textwidth}
    \vspace{1.2cm}
    \includegraphics[width=\textwidth]{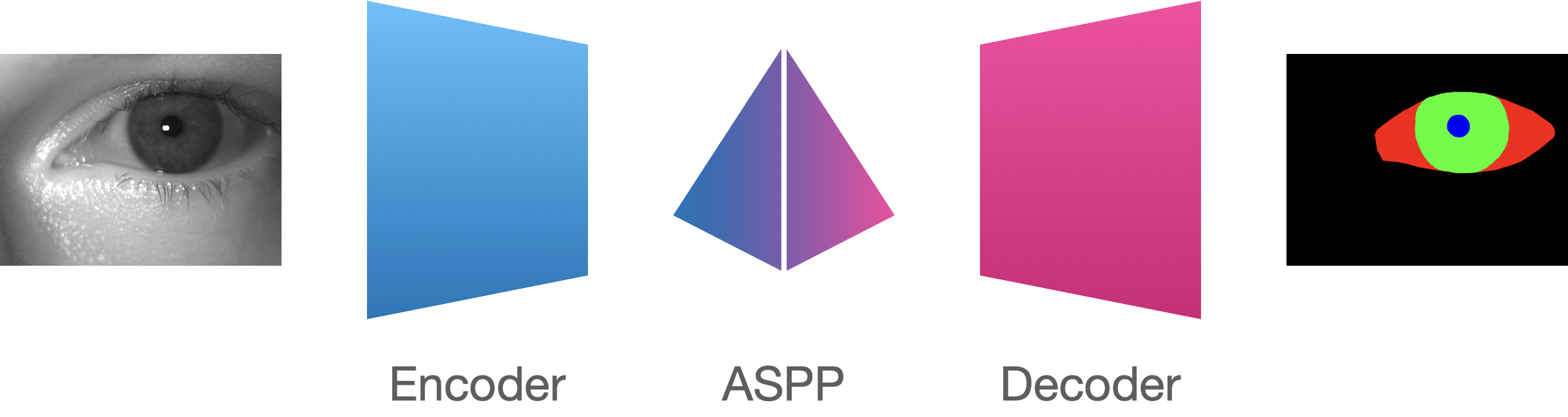}
    \vspace{0.65cm}
    \caption{DeepLabV3 architecture.}
    \label{fig:deeplab}
\end{subfigure}
\hfill
\begin{subfigure}{0.47\textwidth}
    \includegraphics[width=\textwidth]{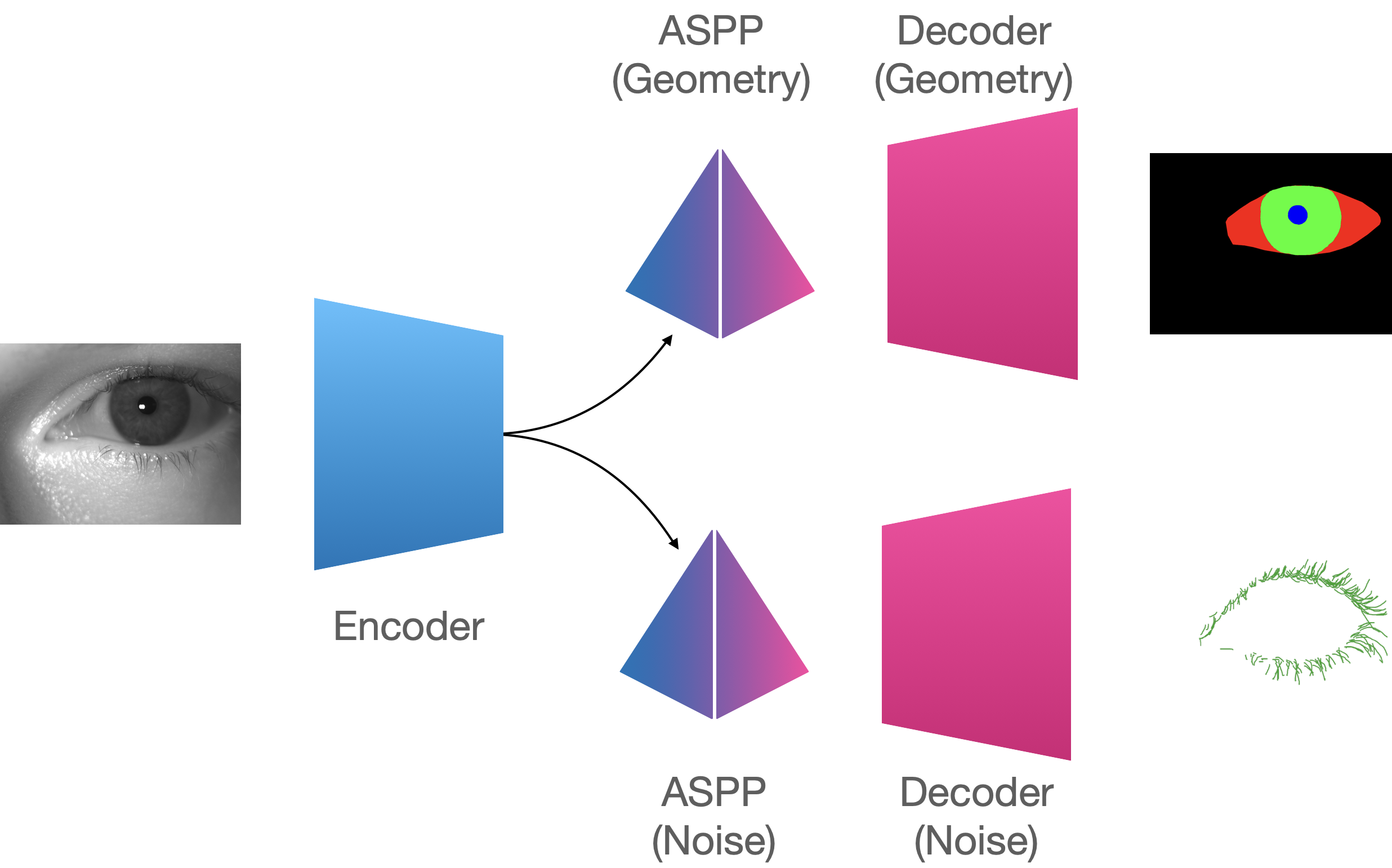}
    \caption{Two-headed architecture.}
    \label{fig:two_headed}
\end{subfigure}
        
\caption{Comparison between the DeepLabV3 \cite{chen2017rethinking} architecture (fig. \ref{fig:two_headed}) and our two-headed approach. Our model shares the same Encoder architecture but is composed of two Decoders: Geometry and Noise. The Geometry Decoder is responsible for segmenting eye components, including the eyeball, iris, and pupil. The Noise Head is returning a binary segmentation mask for eyelashes. }
\label{fig:architecutres}
\end{figure}
In this section, we introduce the architecture of our model. Our model is provided in fig. \ref{fig:two_headed}. The model is based on standard DeepLabV3 \cite{chen2017rethinking} architecture (see fig \ref{fig:deeplab}), which is composed of an Encoding network, Atrous Spatial Pyramid Pooling (ASPP), and  Decoding network. Our approach enriches the existing architecture by providing separate decoding heads composed of dedicated ASPP and Decoder: Geometry and Noise. The geometry head is responsible for segmenting eye subregions, including the eyeball, pupil, and iris. The Noise head returns the binary segmentation mask for eyelashes. 

The motivations behind the two-headed model are as follows. First, the two segmentation tasks are similar, and features jointly extracted in one Encoder should also be useful for both problems. Second, separating heads for the task, instead of taking one decoding path, delivers the possibility of application of convex prior \cite{liu2020convex}, which is applicable for convex shapes aggregated in subregions. Third, the ground truth annotations for eyelashes may not be consistent among the different annotators because they used different annotation markers. Aligning the segmentation masks for eyelashes would be much easier, assuming separate binary annotations. 

We train the entire model in an end-to-end fashion, jointly optimizing the parameters of Encoder and Decoders. Selecting an adequate training objective is one of the critical components while training the model. Therefore, we investigated various types of losses that can be utilized and introduced them in this section.



\textbf{Dice loss.} Let assume, that the output returned by the model for a single pixel at location $(i,j)$ is denoted by $\mathbf{p}_{i,j}$, where $k$-th element $p_{i,j,k}$ stays behind probability of the $k$-th class for a given pixel. We assume that the ground truth class for a corresponding pixel is given as a one-hot coding vector $\mathbf{y}_{i,j}$. The dice loss \cite{sudre2017generalised} is inspired by a similarity measure between two objects, and may be interpreted as the ratio between intersection and sum of two regions. Formally, it is defined as:

\begin{equation}
L_{Dice} = \sum_{k} (1 - \frac{2 \cdot \sum_{j,i} y_{i, j, k} \cdot p_{i,j,k}}{\sum_{j,i}  y_{i,j,k}^2 + \sum_{j,i} p_{i,j,k}^2})
\label{eq:dice}
\end{equation}

\textbf{Boundary loss} Boundary loss \cite{chaudhary2019ritnet} is weighting classical cross-entropy with the coefficient that indicates the boundaries of the segmentation mask:

\begin{equation}
L_{B} = \sum_{j,i} \sum_{k} b_{i,j} \cdot y_{i, j, k} \log p_{i,j,k},
\label{eq:boundary}
\end{equation}
where $b_{i,j}$ is an element of boundary map $\mathbf{b}$, and is equal $1$, if $(i,j)$ is on boundary of segmentation regions. Practically, due to the annotation uncertainty, the Gaussian kernel is applied to blur the map $\mathbf{b}$, or the width of the contour annotation can be increased. 


\textbf{Surface loss} Surface loss \cite{chaudhary2019ritnet} utilizes so-called signed distance function $s_k(\cdot)$:

\begin{equation*}
s_k(\mathbf{x}) = \begin{cases}
-d(\mathbf{x}, \partial\Omega_k) & x \in \Omega_k\\
d(\mathbf{x}, \partial\Omega_k)  & x \notin \Omega_k,
\end{cases}
\end{equation*}
where $\Omega_k$ is the considered region, $\partial \Omega_k$ is boundary of the region, and $d(x, \partial\Omega_k)$ is a distance between the boundary and given point $\mathbf{x}$. In our approach we identify the $4$ regions, such that $\Omega_1$ stays behind \emph{background}, $\Omega_2$ represents \emph{eyeball} region, $\Omega_3$ denotes \emph{pupil} and $\Omega_4$ \emph{iris} areas. We can further construct the tensor $\mathbf{S}$, where element $s_{k}((i,j))=s_{k,i,j}$ represents signed distance function for $k$-th region i point $(i,j)$. 


We define the surface loss in the following manner:

\begin{equation}
L_{S} = \sum_{j,i} \sum_{k} s_{k, i,j} \cdot p_{i,j,k}.
\label{eq:surface}
\end{equation}
Practically, we would like to have $p_{i,j,k}$ equal $1$ for all negative distance values (inside the considered region) and $0$ for points with positive distance values (outside). Moreover, in contrast to the boundary loss, the surface loss is focused more on regions far from segmentation boundaries, eliminating some incorrectly classified additional regions besides the main segmentation area. 

\textbf{Class balancing} The typical criteria used during training can be enriched by incorporating additional, class-dependent weights $c_k$. Practically, the higher weight values are assigned to the less numerous classes in the segmentation map. The class weights can be considered to be a hyperparameter, or can be calculated using the following formula: $c_k = \frac{N}{K \cdot N_k}$, where $\sum_{j,i} y_{i, j, k}=N_k$ is the number of pixels segmented as $k$-th class.  

\textbf{Final loss.} In this work, we consider two separate heads. One of them returns probability maps for eye components (we refer to it as geometry head), and the second one returns binary segmentation for noise artifacts, mainly eyelashes (noise head). The total loss is defined as:

\begin{equation}
L = L_{geo} + \lambda_{noise} \cdot L_{noise},
\label{eq:total}
\end{equation}
where $\lambda_{noise}$ is coefficient that represents the importance of noise coefficient in the aggregated criterion. We define the geometry loss as:

\begin{equation}
L_{geo} = (1 - \lambda_{S}) \cdot L_{Dice_C, geo} +\lambda_{S} \cdot L_{S, geo} + \lambda_{B, geo} \cdot L_{B, geo},     
\label{eq:total_geo}
\end{equation}
where $L_{Dice_C, geo}$ is the class-sensitive Dice loss given by eq. \eqref{eq:dice}, $L_{S, geo}$ and $\lambda_{B, geo}$ are surface and boundary losses given by equations \eqref{eq:surface} and \eqref{eq:boundary} respectively. All of the losses are calculated on the geometry head. $\lambda_{B, geo}$ is the coefficient that represents the importance of boundary loss in aggregated criterion. The coefficient $\lambda_{S}$ represents the trade-off between the Dice and boundary losses and is defined as $\lambda_{S}=\frac{current epoch}{max epochs}$. 


We define the loss for noise head in similar way:$L_{noise} = (1 - \lambda_{S}) \cdot L_{Dice_C, noise} +\lambda_{S} \cdot L_{S, noise}.$ Due to the fact, that noise components are rather thick, and boundaries are equivalent to the segmented objects we do not consider boundary loss in our model.



 \textbf{Segmentation with convex prior} The problem that we tackle assumes segmentation of consistent convex shape. Therefore, we propose to enrich the Geometry head with an additional differentiable component that enforces the convexity of the output shape. We follow the method provided in \cite{liu2020convex} utilizing the procedure that transforms the segmentation map to the convex equivalent. The algorithm is implemented so that it is possible to propagate the gradient in the backward pass, and the model can be trained in an end-to-end fashion. 

\begin{figure}[t!]
     \centering
     \begin{subfigure}[b]{0.3\textwidth}
         \centering
         \includegraphics[width=\textwidth]{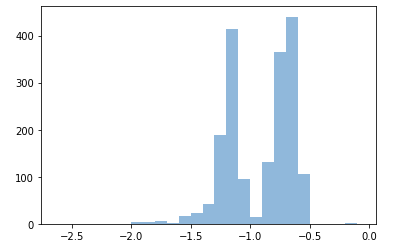}
         \caption{Before alignment.}
         \label{fig:eye_lash_hist_1}
     \end{subfigure}
          \hfill
     \begin{subfigure}[b]{0.3\textwidth}
         \centering
         \includegraphics[width=\textwidth]{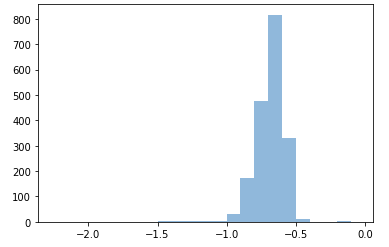}
         \caption{After thinning alignment.}
         \label{fig:eye_lash_hist_2}
     \end{subfigure}
     \hfill
     \begin{subfigure}[b]{0.3\textwidth}
         \centering
         \includegraphics[width=\textwidth]{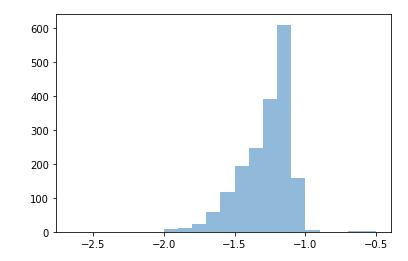}
         \caption{After thickening alignment.}
         \label{fig:eye_lash_hist_3}
     \end{subfigure}
        \caption{Histograms of negative distances.}
        \label{fig:histograms}
\end{figure}

\section{Experiments}

In this section, we describe the results of the experimental evaluation of the proposed model. We perform extensive ablation studies, investigating the impact of various loss combinations on the segmentation quality. 

\textbf{Dataset}. The dataset is composed of $11 560$ training, $ 1448$ validation, and $1420$ test high-resolution near-infrared eye images. Each of the examples has $3$ corresponding geometry segmentation masks (pupil, iris and sclera) annotated by $3$ different annotators and $1$ mask for noise regions. The final mask is obtained by majority voting.


\textbf{Eyelash preprocessing}. The annotations for eye-lashes (noise component) come from Amazon Mechanical Turk and are mainly represented by fragile components marked by lines than sub-region segmentation masks. As a consequence, we distinguish different widths of the lines used for the annotators. In order to achieve a representative dataset for training and evaluation, we propose a novel alignment approach that aims at standardizing the width of segmentation. 

We make use of sign distance function $s_{el}(\mathbf{x})$ that is applied to the eye-lash segmentation map. We calculate the average:
$ \hat{d}_{el} = \frac{1}{N_{el}} \sum_{\mathbf{x} \in \Omega_{el}}-d(\mathbf{x}, \partial\Omega_{el}), 
$
where $N_{el}$ is the number of points inside the region $\Omega_{el}$. In fig. \ref{fig:eye_lash_hist_1} we plot the histogram of $\hat{d}_{el}$ for entire dataset. We can observe two modalities for the histograms, that are associated with two thicknesses of the annotating lines. In order to align the eye-lashes we have to choose one of the strategies: thinning or thickening. Both strategies can be implemented by simple thresholding strategy.  For \emph{thinning strategy},  each element  $x \in \Omega_{el}$, such that $x=1$, and $d(\mathbf{x}, \partial\Omega_{el}) < c$ we set $x=0$. The $c$ is a thresholding value that represents the size of the reduction. Considering \emph{thickening Strategy}, each element  $x \notin \Omega_{el}$, such that $x=0$, and $d(\mathbf{x}, \partial\Omega_{el}) < c$ we set $x=1$. The $c$ is a thresholding value that in this case represents the size of the enlargement.

The histograms after the application of both strategies are provided in fig. \ref{fig:eye_lash_hist_2} and \ref{fig:eye_lash_hist_3}. Further, we consider the thickening strategy, which has no problems with sticky eyelashes and delivers a more balanced segmentation map.  


\textbf{Data prepossessing.} 
The original images were transformed to grayscale and resized from $1440x1080$ to $512x512$. We also applied the set of random augmentations, including cropping, padding, horizontal and vertical flipping, translations, scaling, increasing and decreasing brightness and contrast, adding noise, adding and multiplying by random values the pixels, and blurring. 
\begin{table}[t]
\begin{center}
\begin{scriptsize}
\begin{tabular}{cccccc|cccc}
\textbf{}                            & \multicolumn{5}{c|}{\textbf{Non-Convex}}                                                                                                                         & \multicolumn{4}{c}{\textbf{Convex}}                                                                                                \\ \hline
\multicolumn{1}{c|}{\textbf{Metric}} & \multicolumn{1}{c|}{\textbf{D}} & \multicolumn{1}{c|}{\textbf{D\_b}} & \multicolumn{1}{c|}{\textbf{D+B}}    & \multicolumn{1}{c|}{\textbf{D+S}} & \textbf{D+B+S} & \multicolumn{1}{c|}{\textbf{D+P}}    & \multicolumn{1}{c|}{\textbf{D+P+NE}} & \multicolumn{1}{c|}{\textbf{C}}      & \textbf{D+NE} \\ \hline
\multicolumn{1}{c|}{\textit{Acc}}    & \multicolumn{1}{c|}{0.9895}     & \multicolumn{1}{c|}{0.9884}        & \multicolumn{1}{c|}{\textbf{0.9915}} & \multicolumn{1}{c|}{0.9888}       & 0.9913         & \multicolumn{1}{c|}{0.9872}          & \multicolumn{1}{c|}{0.9885}          & \multicolumn{1}{c|}{0.9874}          & 0.9884        \\ \hline
\multicolumn{1}{c|}{\textit{Prec}}   & \multicolumn{1}{c|}{0.9826}     & \multicolumn{1}{c|}{0.9815}        & \multicolumn{1}{c|}{\textbf{0.9861}} & \multicolumn{1}{c|}{0.9812}       & 0.9855         & \multicolumn{1}{c|}{0.9771}          & \multicolumn{1}{c|}{0.9805}          & \multicolumn{1}{c|}{0.9782}          & 0.9814        \\ \hline
\multicolumn{1}{c|}{\textit{Recall}} & \multicolumn{1}{c|}{0.9831}     & \multicolumn{1}{c|}{0.9821}        & \multicolumn{1}{c|}{\textbf{0.9856}} & \multicolumn{1}{c|}{0.9805}       & 0.9852         & \multicolumn{1}{c|}{0.9832}          & \multicolumn{1}{c|}{0.9824}          & \multicolumn{1}{c|}{0.9826}          & 0.9814        \\ \hline
\multicolumn{1}{c|}{\textit{MIOU}}   & \multicolumn{1}{c|}{0.9697}     & \multicolumn{1}{c|}{0.9676}        & \multicolumn{1}{c|}{\textbf{0.9751}} & \multicolumn{1}{c|}{0.9667}       & 0.9742         & \multicolumn{1}{c|}{0.9652}          & \multicolumn{1}{c|}{0.9675}          & \multicolumn{1}{c|}{0.9655}          & 0.9672        \\ \hline
\multicolumn{1}{l|}{\textit{ICRate}} & \multicolumn{1}{c|}{0.9902}     & \multicolumn{1}{c|}{0.9898}        & \multicolumn{1}{c|}{\textbf{0.9919}} & \multicolumn{1}{c|}{0.9906}       & 0.9918         & \multicolumn{1}{c|}{\textbf{0.9919}} & \multicolumn{1}{c|}{0.9918}          & \multicolumn{1}{c|}{\textbf{0.9919}} & 0.9918       
\end{tabular}
\caption{Ablation studies that show the results for various combinations of the losses for geometry.}
\label{tab:geo}
\end{scriptsize}
\end{center}
\vspace{-1.0cm}
\end{table}


\textbf{Implementation details.} We use the DeeplabV3 architecture with the MobileNet \cite{howard2019searching} as a backbone, and with the number of ASPP features equal $256$. The model is trained using ADAM optimizer, with the polynomial weight scheduler with the learning rate equal $0.0001$. We set $\lambda_{noise}$ equal $3$ to increase the noise component's importance during training.
\textbf{Results.} In this section, we examine the quality of the model considering both noise and geometry heads. The quality of the models is evaluated using standard measures including accuracy (\emph{Acc}), precision (\emph{Prec}), recall (\emph{Recall}) and mean intersection of the union (\emph{MIOU}). In addition, we examine the convexity of iris segmentation that is  represented by the iris convexity rate (\emph{ICRate}) represented by the ratio between the segmented iris area and convex hull of the detected region. First, we evaluate various loss combinations and their impact on the quality of the model including: Dice loss \textbf{(D)}, Dice loss trained with balancing \textbf{(D\_b)}, Boundary loss (\textbf{B}), and Surface loss (\textbf{S}). For convex prior, each of the considered variants is trained with the Dice loss. We consider two scenarios, where the model is trained without enforcing the convexity, and the convex prior is applied as a plugin (\textbf{D+P}), and the variant, where the model is trained together with the prior  (\textbf{C}). For each of the two training variants we consider the variants where the convex prior is not applied to eye-ball (\textbf{NE}). For all of the combinations we assume that noise head is trained using Dice loss.

\begin{wraptable}{r}{5.5cm}
 \vspace{-0.2cm}
\begin{center}
\begin{scriptsize}
\begin{tabular}{c|r|r|r|r}
\textbf{Metric} & \multicolumn{1}{c|}{\textbf{D}} & \multicolumn{1}{c|}{\textbf{D\_b}} & \multicolumn{1}{c|}{\textbf{D+S}} & \multicolumn{1}{c}{\textbf{D\_b+S}} \\ \hline
\textit{Acc}    & \textbf{0.9246}                 & 0.9233                             & 0.9240                            & 0.9230                              \\ \hline
\textit{Prec}   & \textbf{0.7581}                 & 0.7553                             & 0.7563                            & 0.7549                              \\ \hline
\textit{Recall} & 0.8652                          & 0.8668                             & 0.8654                            & \textbf{0.8675}                     \\ \hline
\textit{MIOU}   & \textbf{0.6949}                 & 0.6928                             & 0.6936                            & 0.6926                             
\end{tabular}
\end{scriptsize}
\caption{Ablation studies that show the results for various combinations of the losses for noise head.}
\label{tab:noise}
\end{center}
\end{wraptable}

The results are provided in tab. \ref{tab:geo}. The overall quality of the proposed approach is significantly better than the results reported for previous methods \cite{rot2018deep,liu2020convex}, with the accuracy at level $99\%$ and precision-recall values over $0.98$. It can be noticed that incorporating boundary loss increases the MIOU from $0.9697$ to $0.9751$. Including additional loss components and balancing the class distribution do not increase the quality of the model. The application of convex prior, both as a plugin and during training, increases the convexity rate of the \emph{iris} compared to the scenario where the model is trained only using Dice loss. However, a similar convexity rate can be achieved while using boundary loss. Moreover, due to the fact that the eyeball component is not always convex, enforcing convexity only on the iris and pupil increases the segmentation quality of the model. We also apply ablation studies on the noise head (assuming the best configuration for geometry). We consider the following loss combinations:\textbf{Dice Loss balanced (D\_b)} \textbf{Dice Loss + Surface Loss (D + S)}, and \textbf{Dice Loss balanced + Surface Loss (D\_b + S)}. The results of this experiment are provided tab. \ref{tab:noise}. It can be observed that modifying the loss function associated with the noise head does not have an impact on the quality of the model.

\section{Conclusions}

This paper proposes a novel architecture for eye segmentation tailored for person identification. We examine various types of losses during the training procedure, incorporating additional information about boundaries and signed distance values. We also investigate the possibility of applying the convex prior, in order to enforce the convex shapes of the components. Finally, we evaluate the segmentation model to learn informative representation for identification, assuming a given post-processing pipeline. Future works mainly focus on post-processing steps to increase the quality of the representation for identification. Moreover, we are going to investigate the possibility of learning the representations using an end-to-end approach. 

\newpage 
\bibliographystyle{lnig}
\bibliography{biosig_short}

\end{document}